%% file: 0Paper.tex
\documentclass[letterpaper]{article} 
\usepackage{aaai2026}  
\usepackage{times}  
\usepackage{helvet}  
\usepackage{courier}  
\usepackage[hyphens]{url}  
\usepackage{graphicx} 
\usepackage{booktabs}
\usepackage{color,xspace}
\usepackage{amssymb}
\usepackage{subcaption}
\usepackage{xcolor} 
\usepackage{color} %
\usepackage{natbib}  
\usepackage{caption} 
\usepackage{algorithm}
\usepackage{algorithmic}
\usepackage{amsmath}
\usepackage{newfloat}
\usepackage{listings}
\DeclareCaptionStyle{ruled}{labelfont=normalfont,labelsep=colon,strut=off} 
\floatstyle{ruled}
\newfloat{listing}{tb}{lst}{}
\floatname{listing}{Listing}
\title{Emotion and Intention Guided Multi-Modal Learning for Sticker Response Selection}
\author{
    Yuxuan Hu\textsuperscript{\rm 1,2}\equalcontrib, Jian Chen\textsuperscript{\rm 1,3}\equalcontrib, Yuhao Wang\textsuperscript{\rm 1}, Zixuan Li\textsuperscript{\rm 2}, Jing Xiong\textsuperscript{\rm 3} \\ Pengyue Jia\textsuperscript{\rm 1}\footnotemark[2], Wei Wang\textsuperscript{\rm 2}\thanks{Corresponding authors.}, Chengming Li\textsuperscript{\rm 2}\footnotemark[2], Xiangyu Zhao\textsuperscript{\rm 1}\footnotemark[2]
}
\affiliations{
    \textsuperscript{\rm 1}City University of Hong Kong, Hong Kong, China\\
    \textsuperscript{\rm 2}Shenzhen MSU-BIT University, Shen Zhen, China\\
    \textsuperscript{\rm 3}The University of Hong Kong, Hong Kong, China\\

    yuxuanhu7-c@my.cityu.edu.hk, ccccccj03@connect.hku.hk, licm@smbu.edu.cn, xianzhao@cityu.edu.hk
}

\usepackage{bibentry}

\begin{document}

\maketitle

\begin{abstract}
Stickers are widely used in online communication to convey emotions and implicit intentions. The Sticker Response Selection (SRS) task aims to select the most contextually appropriate sticker based on the dialogue. However, existing methods typically rely on semantic matching and model emotional and intentional cues separately, which can lead to mismatches when emotions and intentions are misaligned. To address this issue, we propose \textbf{E}motion and \textbf{I}ntention \textbf{G}uided \textbf{M}ulti-Modal \textbf{L}earning (\textbf{EIGML}). This framework is the first to jointly model emotion and intention, effectively reducing the bias caused by isolated modeling and significantly improving selection accuracy. Specifically, we introduce Dual-Level Contrastive Framework to perform both intra-modality and inter-modality alignment, ensuring consistent representation of emotional and intentional features within and across modalities. In addition, we design an Intention-Emotion Guided Multi-Modal Fusion module that integrates emotional and intentional information progressively through three components: Emotion-Guided Intention Knowledge Selection, Intention-Emotion Guided Attention Fusion, and Similarity-Adjusted Matching Mechanism. This design injects rich, effective information into the model and enables a deeper understanding of the dialogue, ultimately enhancing sticker selection performance. Experimental results on two public SRS datasets show that EIGML consistently outperforms state-of-the-art baselines, achieving higher accuracy and a better understanding of emotional and intentional features. The code is released at https://github.com/Applied-Machine-Learning-Lab/EIGML.
\end{abstract}

\input{1Introduction}

\input{2RelatedWork}
\input{3Framework}
\input{4Experiments}

\input{5Conclusions}
\section{Acknowledgments}
This research was partially supported by National Natural Science Foundation of China (No.62502404), Hong Kong Research Grants Council (Research Impact Fund No.R1015-23, Collaborative Research Fund No.C1043-24GF, General Research Fund No.11218325), Institute of Digital Medicine of City University of Hong Kong (No.9229503), Huawei (Huawei Innovation Research Program), Tencent (CCF-Tencent Open Fund, Tencent Rhino-Bird Focused Research Program), Alibaba (CCF-Alimama Tech Kangaroo Fund No. 2024002), Didi (CCF-Didi Gaia Scholars Research Fund), Kuaishou, and Bytedance, Innovation Team Project of Guangdong Province of China (No. 2024KCXTD017), Shenzhen Science and Technology Foundation (No. JCYJ20240813145816022).

\appendix
\nocite{*}

\bibliography{6Reference}



\end{document}

%% file: 1Introduction.tex
\section{Introduction}

\begin{figure}[t]
\setlength\abovecaptionskip{0.2\baselineskip}
\setlength\belowcaptionskip{0.2\baselineskip}
    \centering
    \includegraphics[width=\linewidth]{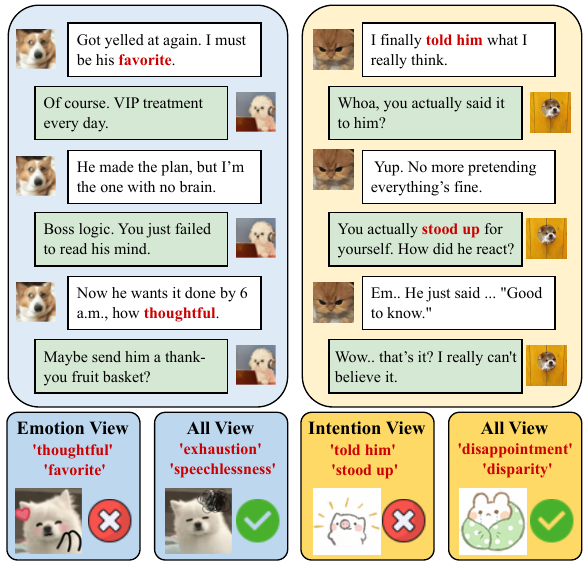}
    \caption{Examples for potential failures in sticker selection
when relying solely on emotional or intentional cues.
    }
    \label{fig:introduction}
    \vspace{-20pt} 
\end{figure}


With the rapid proliferation of social media platforms, stickers have attracted significant attention due to their unique ability to express emotions and convey communicative intent~\cite{zhao2021affective,zhang2024stickerconv}. An appropriately selected sticker can not only effectively highlight the emotions and intentions which users wish to express, but also serve as a substitute for complex textual responses~\cite{chen2024tgca, tang2019emoticon}. Consequently, the task of Sticker Response Selection (SRS) has emerged as an active research area and garnered significant attention.

Compared to typical image-text matching tasks~\cite{zhang2022negative, wu2019learning,zhang2023skeletal}, SRS poses more complex, highly context-sensitive challenges. Stickers serve not only as visual symbols but also as rich multi-modal carriers of emotional and intentional cues~\cite{wang2024towards, chen2025mghftmultigranularityhierarchicalfusion, shi2024integrating}, whose meanings vary greatly across conversational contexts. Thus, models must move beyond surface semantic alignment to effectively capture emotional signals and infer communicative intent for deeper multi-modal understanding~\cite{chee2025persrv, ge2022towards, liu2022ser30k}. However, many methods still rely on shallow semantic matching~\cite{gao2020learning, fei2021towards}, failing to grasp subtle emotional and intentional cues. In real dialogues, stickers convey not only emotions but also tone, social stance, and complex communicative strategies. Ignoring these leads to semantically plausible yet contextually inappropriate choices.

Although some studies incorporate emotional or intentional information to improve multi-modal alignment~\cite{xia2024perceive, wang2025new}, they often treat the two separately, overlooking their intertwined nature in human communication. This separation may cause over-reliance on one aspect and neglect of the other, leading to mismatches in nuanced sticker selection. As shown in Figure~\ref{fig:introduction}, models focusing only on emotion may miss subtleties like sarcasm or irony, while those centered on intention may ignore emotional undertones, resulting in contextually inappropriate selections. These cases show that isolating emotion or intention is insufficient and even harm overall performance. A unified emotion-intention framework is thus needed to better capture the contextual communicative functions of stickers.

To address these challenges, we propose the Emotion and Intention Guided Multi-Modal Learning (EIGML) framework to enhance SRS accuracy by jointly modeling emotional and intentional cues. The framework has two core modules targeting key issues. First, to overcome the limitations of semantic matching in capturing subtle emotional variations and implicit intentions, we introduce the Dual-Level Contrastive Framework (DLCF), which boosts sensitivity to emotional and intentional features via intra-modality and inter-modality alignment. Second, to mitigate the bias caused by modeling emotion and intention in isolation, we design the Intention-Emotion Guided Multi-Modal Fusion (IEGMF) mechanism, comprising Emotion-Guided Intention Knowledge Selection (EIKS), Intention-Emotion Guided Attention Fusion (IEGA), and Similarity-Adjusted Matching Mechanism (SAMM). Together, they inject rich, effective information to deepen contextual understanding and improve selection. Extensive experiments on two large-scale datasets, StickerChat and DSTC10-MOD, demonstrate that our method consistently outperforms state-of-the-art models across multiple metrics, proving its effectiveness and robustness in complex scenarios. Our contributions include:



\begin{itemize}
\item To the best of our knowledge, this is the first framework to jointly model emotion and intention explicitly for the task of Sticker Response
Selection (SRS) task, reducing mismatches caused by isolated modeling and improving sticker selection accuracy in dialogue. We also identify the critical challenges in existing SRS methods rendered by the emotional and intentional cues, which have been less explored in the literature.


\item We propose a novel EIGML framework, which consists of two core components designed to address the mismatch between emotional and intentional features across visual and textual modalities. Building upon semantic alignment, EIGML introduces cross-modal alignment of emotion and intention representations to enhance overall modality consistency. Moreover, EIGML employs a three-stage progressive fusion strategy, where emotional and intentional cues are gradually injected to guide the integration of visual and textual features, thereby improving the effectiveness of sticker selection.


\item 
Extensive experimental results on two public benchmark datasets validate the effectiveness and efficiency of the proposed EIGML framework.
\end{itemize}

%% file: 2RelatedWork.tex
\section{Related Work}
Existing sticker response selection models can be grouped into shallow semantic matching and disjoint emotion–intention modeling.

\subsubsection{Shallow Semantic Matching Methods.}

Early approaches, such as~\citet{laddha2020understanding}, use clustering-based methods to predict the next message, which is then replaced with a sticker. Later, a deep interaction network is proposed in~\citet{gao2020learning}, which leverages cross-attention to extract multi-modal features for matching-based sticker retrieval. In~\citet{fei2021towards}, both text and sticker prediction are framed as sequence generation tasks using a unified framework with pretrained GPT-2~\cite{radford2019language} to jointly encode dialogue context for sticker selection. A multi-task learning framework is introduced by~\citet{zhang2022selecting} to improve understanding of both sticker and dialogue semantics. While these methods show promising semantic matching results, they often neglect the rich emotional and intentional information in stickers, underscoring the need for explicit modeling of such information.
\subsubsection{Separate Emotion and Intention Modeling Methods.} 
In recent years, several studies have attempted to go beyond semantic matching by incorporating emotional or intentional understanding of both stickers and dialogue. CKS~\cite{chen2024deconfounded} integrates commonsense knowledge to better recognize emotional expressions and extract unbiased visual features, thereby improving alignment between stickers and dialogues. PBR~\cite{xia2024perceive} focuses on emotional features within both stickers and dialogues, enabling emotion-aware sticker selection. Some works~\cite{wang2025new,liang2025reply} construct datasets with intention labels to incorporate intention into sticker response selection. While these methods introduce emotional or intentional signals, they typically model them separately, failing to capture their intertwined nature. To address this limitation, we propose a unified framework that integrates emotion and intention across modalities, improving both the accuracy and contextual relevance of sticker response selection.

%% file: 3Framework.tex
\section{Method}

\subsection{Overview}
We first introduce the problem definition of SRS, followed by an overview of our proposed framework EIGML.

\begin{figure*}
\setlength\abovecaptionskip{0.2\baselineskip}
\setlength\belowcaptionskip{0.2\baselineskip}
    \centering
    \includegraphics[width=0.925\linewidth]{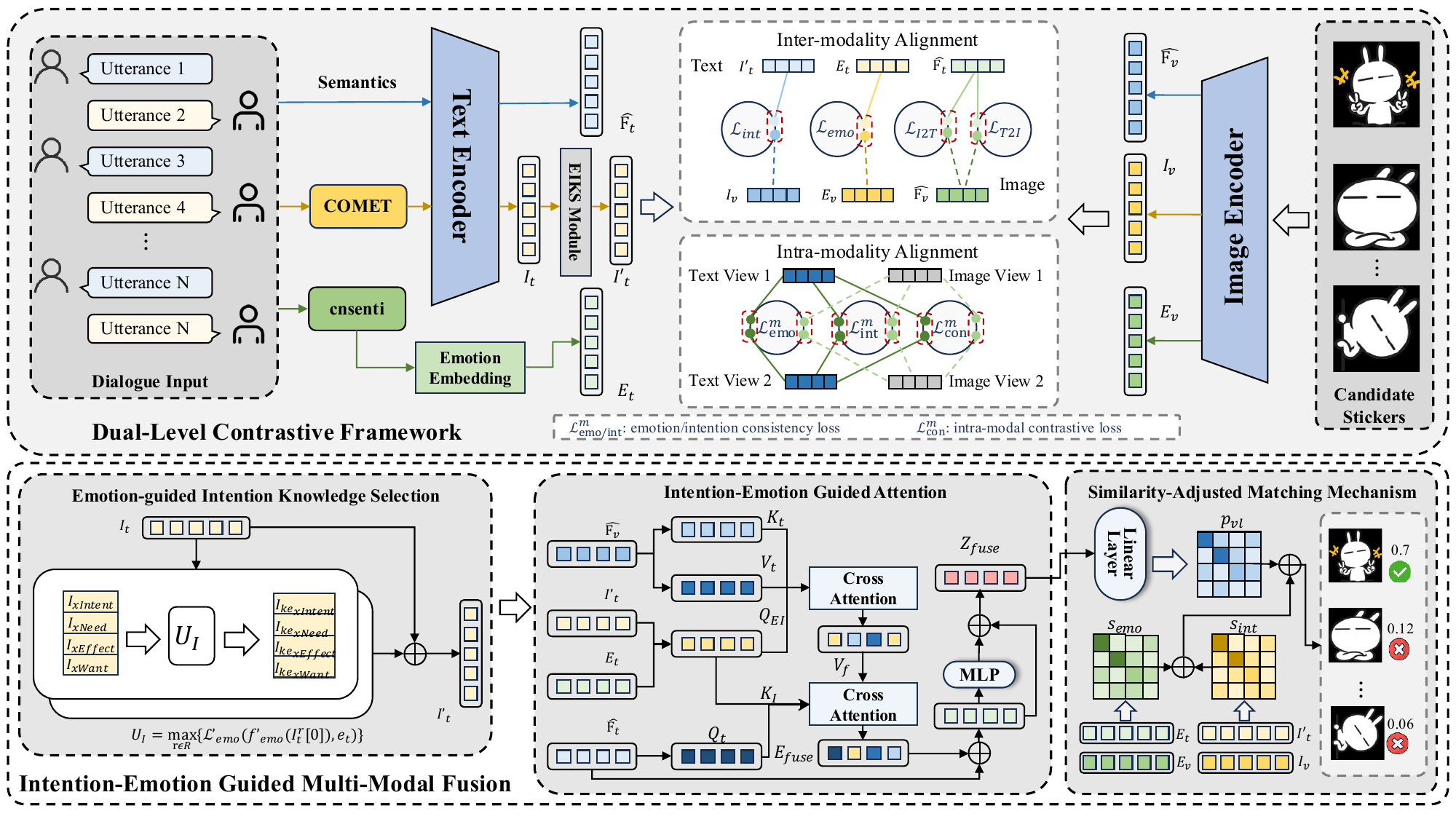}
    \caption{Overview of our proposed approach. Dual-Level Contrastive Framework enhances intra-modality and inter-modality alignment, while Intention-Emotion Guided Multi-Modal Fusion integrates three submodules: (1) Intention-Emotion Attention Fusion, (2) Emotion-Guided Intention Selection, and (3) Similarity-Adjusted Matching Mechanism.}
    \label{fig:framework}
    \vspace{-15pt} 
\end{figure*}


\subsubsection{Problem Definition.}
Given a multi-turn dialogue history $U=\{u_{1}, u_{2}, ...,u_{N_U}\}$ and a set of candidate stickers $S=\{s_{1}, s_{2}, ...,s_{N_S}\}$, where $N_U$ and $N_S$ denote the number of utterances and candidate stickers respectively, the SRS task aims to select the most appropriate sticker by understanding various cues in the dialogue.
Specifically, we aim to leverage semantic, emotional, and intentional information from the text to train a model that predicts the suitability of each candidate sticker. The training objective minimizes the binary cross-entropy (BCE) loss between the predicted matching scores and ground-truth labels defined as:
\begin{equation}
\min_{\theta} \mathcal{L}_{ma} = -\frac{1}{N} \sum_{i=1}^{N} \text{BCE} \left( f_\theta (U_i, S_i), y_i \right),
\label{equ:obj}
\end{equation}
where $\theta$ represents the trainable parameters of the model, $N$ is the total number of samples in the dataset, and $y_i=\{y^{i}_1,y^{i}_2,...,y^{i}_{N_{s}}\}$ indicates whether the candidate stickers in the $i$-th sample matches the dialogue information $U_{i}$. 


\subsubsection{Framework Overview.}

To overcome the limitations of purely semantic matching in capturing subtle emotions and implicit intentions, as well as the mismatches from modeling them separately, we propose the Emotion and Intention Guided Multi-Modal Learning (EIGML) framework. As shown in Figure~\ref{fig:framework}, EIGML comprises two core components: Dual-Level Contrastive Framework (DLCF) and Intention-Emotion Guided Multi-Modal Fusion (IEGMF).
DLCF enhances the representation of emotional and intentional cues and improves multi-modal alignment via both intra-modality and inter-modality strategies. IEGMF integrates Emotion-Guided Intention Knowledge Selection (EIKS), Intention-Emotion Guided Attention Fusion (IEGA), and Similarity-Adjusted Matching Mechanism (SAMM) to enable deep visual-textual fusion, effectively capturing the complementarity between emotion and intention.

\subsection{Dual-Level Contrastive Framework}


Although emotional and intentional cues are crucial for SRS~\cite{xia2024perceive, wang2025new}, existing methods typically model them independently, often resulting in semantic mismatches. To bridge this gap, we propose the first unified framework that jointly models emotion and intention to address their misalignment. A core component, DLCF, performs inter-modality alignment to ensure semantic consistency across modalities and intra-modality alignment to enhance the representational capacity of each modality. This design significantly boosts representation quality and reduces mismatches in sticker selection.


\subsubsection{Inter-Modality Alignment.}

Conventional semantic alignment~\cite{gao2020learning,fei2021towards} falls short in capturing the subtle emotional–intentional interplay across modalities. To address this, we propose a cross-modal alignment mechanism that explicitly targets emotion and intention. This design enhances the understanding of emotional and intentional cues and semantic consistency between modalities, yielding significantly improved matching performance.

Image and text data are passed through the visual encoder and the textual encoder to get features $F_{v}$ and $F_{t}$, respectively. We adopt the InfoNCE loss~\cite{oord2018representation} for image-text semantic alignment. The \texttt{[CLS]} token features of images and texts are projected into a shared embedding space via linear mappings: $\hat{F}_{v} = F^{\text{cls}}_{v} W_{v}$, $\hat{F}_{t} = F^{\text{cls}}_{t} W_{t}$. In image-to-text alignment, a sticker and its paired dialogue form a positive pair, with other dialogues as negatives; in text-to-image alignment, the dialogue and its matched sticker form a positive pair, with other stickers as negatives. The losses are defined as follows:
\begin{equation}
    \mathcal{L}_{I2T} = -\frac{1}{K} \sum_{i=1}^K \log \frac{\exp(\hat{F}_{v,pos}^i \cdot \hat{F}_t^i/\tau)}{\sum_{j=1}^K \exp(\hat{F}_{v,pos}^i \cdot \hat{F}_t^j/\tau)},
    \label{equ:i2tse}
\end{equation}
\begin{equation}
    \mathcal{L}_{T2I} = -\frac{1}{K} \sum_{i=1}^K \log \frac{\exp(\hat{F}_t^i \cdot \hat{F}_{v,pos}^i/\tau)}{\sum_{j=1}^K \exp(\hat{F}_t^i \cdot \hat{F}_{v,neg}^j/\tau)},
    \label{equ:t2ise}
\end{equation}
where $K$ is the batch size, $\hat{F}_{v,pos}^i$ and $\hat{F}_{v,neg}^i$ denote the visual embedding of the positive and negative stickers for the $i$-th dialogue.

For each dialogue, we use the cnsenti~\cite{Deng2019} toolkit to derive an emotion distribution $\mathbf{e}_t$ over seven categories. The most prominent is embedded via the emotion embedding matrix $E_{\text{emo}}$ to obtain the textual emotion embedding $E_t$. For the image, a linear head $f_{\text{emo}}$ maps the visual \texttt{[CLS]} feature $F_i^{\text{cls}}$ to its emotion representation $E_v$, defined as:

\begin{equation}
    E_t = E_{\text{emo}}\left[\arg\max(\mathbf{e}_t)\right], \quad
    E_v = f_{\text{emo}}(F_v^{\text{cls}}).
    \label{equ:emo_fea}
\end{equation}

To achieve cross-modal emotion alignment, we minimize the Kullback–Leibler (KL) divergence between the predicted emotion distribution of the matched image $E^{\text{pos}}_v$ and the corresponding text $E_t$ defined as follows: 
\begin{equation}
    \mathcal{L}_{\text{emo}} = KL(E^{\text{pos}}_v,E_t),
    \label{equ:emo_align}
\end{equation}
where $E^{\text{pos}}_v$ means the feature of the correct sticker. 

Next, we leverage the generative commonsense transformer model COMET~\cite{bosselut2019comet} for commonsense inferences from the input text across four relations $R=\{$\textit{xIntent}, \textit{xNeed}, \textit{xWant}, \textit{xEffect}$\}$. These inferences are concatenated as $C_{t}$, defined as:
\begin{equation}
C_r = \text{COMET}(T, r), \quad C_t = \oplus_{r \in R} C_r,
\label{equ:comet}
\end{equation}
where $r \in R$ denotes the relation type, and $\oplus$ is the concatenation operation. 

The commonsense inference $C_t$ is encoded via the text encoder to obtain the initial intention embedding $I_t = \mathcal{F}t(C_t)$, which is further refined by the EIKS module into $I'_t$ (detailed in the next section). For the visual modality, a projection head $f_{\text{int}}$ maps the sticker’s \texttt{[CLS]} feature $F_i^{\text{cls}}$ to its intention embedding $I_v = f_{\text{int}}(F_i^{\text{cls}})$. The intention alignment loss $\mathcal{L}_{\text{int}}$, based on InfoNCE (Equation~\eqref{equ:i2tse}, \eqref{equ:t2ise}), encourages consistency between $I_v$ and $I'_t$. The total inter-modality loss is defined as:
\begin{equation}
    \mathcal{L}_{inter} = \mathcal{L}_{I2T} + \mathcal{L}_{T2I} + w_{e}\cdot\mathcal{L}_{\text{emo}} + w_{i}\cdot\mathcal{L}_{\text{int}}.
    \label{equ:inter_loss}
\end{equation}
where $w_{e}$ and $w_{i}$ are the weights for $\mathcal{L}_{\text{emo}}$ and $\mathcal{L}_{\text{int}}$.


\subsubsection{Intra-Modality Alignment.} While cross-modal alignment ensures inter-modality consistency, strong intra-modality representations are crucial. Emotional and intentional cues often lie within each modality and may be missed by shallow encoders. To address this, we introduce intra-modality alignment to enhance each modality’s expressiveness and discriminability, reinforcing emotion and intention understanding to support improved cross-modal matching.

Specifically, we perform intra-modality alignment by generating two augmented views via independent dropout and separate projection heads~\cite{gao2021simcse}. These views are mapped to emotion and intention spaces using the shared intention head for inter-modality alignment. For each modality $m \in \{v,t\}$, the emotion and intention consistency losses are  symmetric KL divergence, which are defined as:
\begin{equation}
    \mathcal{L}^{m}_{*} = \frac{1}{2} \left[ KL(p^{m}_{1,*} \| p^{m}_{2,*}) + KL(p^{m}_{2,*} \| p^{m}_{1,*}) \right],
    \label{equ:sslei}
\end{equation}
where $p^{m}_{i,*}$ is the softmax-normalized prediction from the $i$-th view and $* \in \{\text{emo}, \text{int}\}$.
Additionally, we employ InfoNCE-style instance-level contrastive learning between the two views using cosine similarity. The contrastive loss for modality $m$ is defined as:
\begin{equation}
    \mathcal{L}_{con}^{m}=\frac{1}{2}[\mathcal{L}_{NCE}(z^{m}_{1},z^{m}_{2})+\mathcal{L}_{NCE}(z^{m}_{2},z^{m}_{1})],
    \label{equ:ssla}
\end{equation}
where $z_{1},z_{2}$ are are normalized features from two augmented views. The final intra-modality loss is the sum of emotion KL losses, intention KL losses, and the instance contrastive loss are defined as:
\begin{equation}
    \mathcal{L}_{intra} = \sum_{m\in\{v,t\}}(w_{e}^{m}\cdot\mathcal{L}^{m}_{emo}+w_{i}^{m}\cdot\mathcal{L}^{m}_{int}+\mathcal{L}_{con}^{m}).
    \label{equ:intra}
\end{equation}
where $w_{e}^{m}$ and $w_{i}^{m}$ are the weights for $\mathcal{L}^{m}_{emo}$ and $\mathcal{L}^{m}_{int}$.

\subsection{Intention-Emotion Guided Multi-Modal Fusion}



While emotion and intention are inherently intertwined in real-world communication, existing approaches~\cite{xia2024perceive, wang2025new} often model them in isolation, which may lead to mismatches and misunderstandings in SRS. To address this challenge, we propose Intention-Emotion Guided Multi-Modal Fusion (IEGMF). This module enhances the deep fusion of emotional and intentional knowledge across modalities through three coordinated components: (1) EIKS leverages emotional cues to retrieve relevant intentional knowledge; (2) IEGA enables fine-grained integration of multi-modal features with joint intentional-emotional guidance; and (3) SAMM refines the final decision by correcting for potential misalignment between emotion and intention.

\subsubsection{Emotion-Guided Intention Knowledge Selection.}


To better align intentional knowledge with emotional state and dialogue context, we propose Emotion-Guided Intention Knowledge Selection (EIKS) inspired by~\citet{cai2023improving}. Specifically, a linear classifier $ f^{\prime}_{emo}$ is applied to the text representation $F^{cls}_{t}$ to predict an emotion distribution via softmax. Cross-entropy loss optimizes both the predicted distribution and learned emotion representation, defined as:
 $\mathcal{L}^{\prime}_{emo}=-\sum_{c=1}^{C} e_t \log \sigma(f^{\prime}_{emo}(F^{cls}_{t})),$
where C is the number of emotion categories and $\sigma$ denotes the softmax function.
Then, $f^{\prime}_{emo}$ eliminates irrelevant intentional knowledge by iteratively refining knowledge representations using gradients from the emotion classification task. At each step, the knowledge vector with the highest emotional classification loss is identified as the most irrelevant candidate, defined as:
\begin{equation}
    U_I=\max_{r \in R} \left\{\mathcal{L}^{\prime}_{emo}(f^{\prime}_{emo}(I_t^r[0]), e_t)\right\}.
    \label{equ:k_select}
\end{equation}

We further employ a nonlinear regression method to model the influence of knowledge exclusion, yielding a guidance matrix $G = \nabla_\theta f$. Next, we utilize this matrix to obtain the adjustment vector $\delta$.
This vector is then injected into all knowledge representations to construct knowledge-enhanced representations: $I_{ke}=I_t+\delta$. Following knowledge enhancement, a dynamic gating mechanism $f_{gate}$ is used to adaptively fuse the enhanced knowledge $I_{ke}$ with the knowledge $I_t$ to get $I'_t$, defined as follows:
\begin{align}
&w_{k} = \text{Sigmoid}\left(f_{gate}([I_t; I_{ke}])\right), \\
&I'_t = w_{k} \odot I_{ke} + (1 - w_{k}) \odot I_t,
\label{equ:I't}
\end{align}
where $\odot$ denotes element-wise multiplication.

\subsubsection{Intention-Emotion Guided Attention Fusion.}

To enable fine-grained contextual-level alignment, we design Intention-Emotion Guided Attention Fusion (IEGA) that integrates selected emotional and intentional cues into the multi-modal fusion process, enhancing the joint understanding of dialogue and sticker semantics. Specifically, we first concatenate the intentional knowledge features $I'_{t}$ with the textual emotion representation $E_t$ as the intention-emotion features $E_{IT}$. Given image features $F_{v}$, textual features $F_{t}$, and the Intention-Emotion features $E_{IT}$, we first compute their linear projections: query vectors $Q_{T}$ and $Q_{EI}$ from $F_{t}$ and $E_{IT}$, and key/value vectors $K_{V}$ and $V_{V}$  from $F_{v}$. The query from Intention-Emotion features attends to the image keys and values, generating Intention-Emotion knowledge-aware visual value $V_{EIV}$, defined as:
\begin{equation}
    V_{EIV} = \sigma\left(\frac{Q_{EI} K_{V}^\top}{\sqrt{d_{head}}}\right) V_{V}.
    \label{equ:IEGA1}
\end{equation} 

Then, textual queries attend to the intention-emotion knowledge-aware visual features as:
\begin{equation}
    E_{fuse} =  \sigma\left(\frac{Q_T Q_{EI}^\top}{\sqrt{d_{head}}}\right) V_{EIV},
    \label{equ:IEGA2}
\end{equation}
where $d_{head}$ is the attention head dimension. An MLP $f_{fuse}$ is used for the final output, defined as:
\begin{equation}
    Z_{fuse} = f_{fuse}(\texttt{LN}\left(F_{t} + \texttt{SD} (E_{fuse})\right),    \label{equ:IEGA3}
\end{equation}
where Layer normalization ($\texttt{LN}$), dropout with stochastic depth ($\texttt{SD}$) are applied to improve training stability.
\vspace{-5pt}

\subsubsection{Similarity-Adjusted Matching Mechanism.}
To improve the model’s ability to identify the most appropriate sticker from emotional and intentional perspectives, we propose Similarity-Adjusted Matching Mechanism (SAMM), which refines the matching score by integrating semantic similarity in both the emotional and intentional representation spaces. We first use a linear layer to predict the semantic match score $p_{vl}$ based on $Z_{fuse}$.   Specifically, given the textual emotion and intention embeddings $E_{t}, I'_{t}$ from the dialogue, and their corresponding sticker-level prototypes $E_{v}, I_{v}$, we compute normalized cosine similarities as:
\begin{equation}
s_{\text{emo}} = \frac{1 + \cos(E_{t}, E_{v})}{2}, \quad
s_{\text{int}} = \frac{1 + \cos(I'_{t}, I_{v})}{2}.
\label{equ:s_emo_int}
\end{equation}

These scores are fused into a single similarity score with a learnable parameter $\alpha$ as:
\begin{equation}
s_{\text{EI}} = \alpha \cdot s_{\text{emo}} + (1 - \alpha) \cdot s_{\text{int}},
\label{equ:s_EI}
\end{equation}

We then combine $s_{\text{EI}}$ with the original vision-language matching probability $p_{\text{vl}}$ to obtain the final relevance score with a learnable parameter $\beta$ to balance the contribution of matching confidence and emotion-intention alignment between the vision and text modalities, defined as:
\begin{equation}
p_{\text{final}} = \beta \cdot p_{\text{vl}} + (1 - \beta) \cdot s_{\text{EI}},
\label{equ:p_final}
\end{equation}

\subsection{Training \& Inference}

We train our model using the full combination of losses in an end-to-end way. The whole loss used for training our model can be summarized as:
\begin{equation}
    \mathcal{L}_{total} = \mathcal{L}_{\text{itm}} + \mathcal{L}_{\text{inter}} + \mathcal{L}_{\text{intra}} + w^{\prime}_{e}\cdot\mathcal{L}^{\prime}_{\text{emo}}.
    \label{equ:totalLoss}
\end{equation}
where $w^{\prime}_{e}$ is the weight of loss $\mathcal{L}^{\prime}_{\text{emo}}$.

During inference, we discard all contrastive components and directly use $p_{final}$ as the predicted score. The sticker with the highest score is regarded as the selected response.

%% file: 4Experiments.tex
\section{Experiments}

\subsection{Dataset and Metrics}

\subsubsection{Datasets}

We conduct experiments on two public datasets: StickerChat~\cite{gao2020learning} and the Chinese version of DSTC10-MOD~\cite{fei2021towards}, with detailed statistics shown in Table~\ref{tab:statisticks}. StickerChat contains 320,168 dialogue-sticker pairs for training and 10,000 each for validation and testing. DSTC10-MOD includes 45,000 dialogues and 307 stickers; since the test set is unavailable, we follow~\citet{zhang2022selecting} and evaluate on the validation set. Dialogues are split into multiple samples using the same preprocessing, resulting in 211,575 training and 3,542 test pairs.

\begin{table}[t]
\setlength\abovecaptionskip{0.2\baselineskip}
\setlength\belowcaptionskip{0.2\baselineskip}
  \centering
  
    \begin{tabular}{lccc}
    \toprule
          & Stickers & Dialogue-Sticker Pairs \\
    \midrule
    StickerChat  & 174,695  &  350,168   \\
    DSTC10-MOD &  307  & 215,117 \\
    \bottomrule
    \end{tabular}%
    \caption{Statistics of StickerChat and DSTC10-MODzhe.}
    \vspace{-15pt} 
  \label{tab:statisticks}%
\end{table}%


\subsubsection{Metrics}

Following prior work~\cite{gao2020learning,xia2024perceive}, we adopt Mean Average Precision (MAP) and Recall at position K among 10 candidates ($R_{10}$@K) to evaluate sticker selection performance. MAP reflects the overall ranking quality, while $R_{10}$@K (K = 1, 2, 5) measures the proportion of cases where the correct sticker appears in the top K, indicating retrieval effectiveness at different cutoffs. All results are reported as percentages.
\vspace{-5pt}

\subsection{Implementation Details}

Following~\citet{xia2024perceive}, in StickerChat, each dialogue is paired with one correct sticker and other theme-based distractors, while in DSTC10-MOD, distractors are randomly sampled from the full sticker set. 
We use the pre-trained bert-base-chinese~\cite{devlin2019bert} (max length 512) as the text encoder and ViT-B/16 from ALBEF~\cite{li2021align} as the visual encoder, both producing 768-dimensional features. The model is optimized with AdamW (lr=5e-5) for 5 epochs, with a random seed of 42. Experiments run on 4 NVIDIA A800 GPUs using PyTorch with a batch size of 16 per GPU. Input images are randomly cropped and resized to 128×128. Emotion-related weights ($w_e$, $w_e'$, $w_e^t$, $w_e^v$) and intention-related weights ($w_i$, $w_i^t$, $w_i^v$) are set to 0.5.

\subsection{Comparison with Baselines}
\subsubsection{Baselines.} On the StickerChat dataset, we compare our proposed method, EIGML, with previous methods including PSAC~\cite{li2019beyond}, LSTUR~\cite{an2019neural}, Synergistic~\cite{guo2019image}, SRS~\cite{gao2020learning}, CLIP~\cite{radford2021learning}, ALBEF~\cite{li2021align}, and PBR~\cite{xia2024perceive}. For the DSTC10-MOD dataset, we compare EIGML with SRS, MOD-GPT~\cite{fei2021towards}, CLIP, MMBERT~\cite{zhang2022selecting}, and PBR. 

Table~\ref{tab:stickerchat} shows the comparison results on the StickerChat dataset. Compared to visual question answering methods (PSAC and Synergistic), recommendation methods (LSTUR), pre-trained multi-modal retrieval models (CLIP and ALBEF), and even SRS task-specific methods (SRS and PBR), our proposed EIGML achieves the best performance. As shown in Table~\ref{tab:stickerchat}, the best performance of the previous methods is PBR, which achieves 79.2\% on MAP and 69.3\% on $R_{10}@1$. By contrast, our EIGML selects stickers that align with both the emotional and intentional requirements of the dialogue while considering the semantics alignment, thus outperforming previous methods with 2\% on MAP and 3\% on $R_{10}@1$, which suggests that the joint modeling of emotional and intentional cues is effective in enhancing SRS accuracy and can partially mitigate mismatches that arise when emotions and intentions are misaligned.
\begin{table}[t]
\setlength\abovecaptionskip{0.2\baselineskip}
\setlength\belowcaptionskip{0.2\baselineskip}
  \centering
  
    \begin{tabular}{lcccc}
    \toprule
          & MAP   & $R_{10}@1$ & $R_{10}@2$ & $R_{10}@5$ \\
    \midrule
    PSAC  & 66.2  & 53.3  & 64.1  & 83.6  \\
    LSTUR & 68.9  & 55.8  & 68.0  & 87.4  \\
    Synergistic & 59.3  & 43.8  & 56.9  & 79.8  \\
    SRS   & 70.9  & 59.0  & 70.3  & 87.2  \\
    CLIP  & 70.9  & 59.1  & 70.3  & 86.8  \\
    ALBEF & 76.8  & 67.0  & 75.6  & 90.0  \\
    PBR  & \underline{79.2}  & \underline{69.3}  & \underline{79.5}  & \underline{93.5}  \\
    EIGML  &   \textbf{81.2{\large *}}    &   \textbf{72.3{\large *}}    &  \textbf{81.3{\large *}}     & \textbf{93.9{\large *}} \\
    \bottomrule
    \end{tabular}%
    \caption{Evaluation results on the StickerChat dataset, best results in \textbf{bold} while second with \underline{underline}. "\textbf{{\large *}}'' indicates the statistically significant improvements (i.e., two-sided t-test with $p<0.05$) over the best baseline.}
  \label{tab:stickerchat}%
  \vspace{-10pt}
\end{table}%

\begin{table}[t]
\setlength\abovecaptionskip{0.2\baselineskip}
\setlength\belowcaptionskip{0.2\baselineskip}
  \centering
  
    \begin{tabular}{lrrrr}
    \toprule
       & MAP & $R_{10}@1$ & $R_{10}@2$ & $R_{10}@5$ \\
    \midrule
    SRS   &    50.3   &  30.5     &    54.2   & 71.3 \\
    MOD-GPT &   52.3    &  31.2     &    54.8   & 72.1 \\
    CLIP  &     54.9  &   38.4    &    56.5   & 52.3 \\
    MMBERT &   57.7    &   37.1    &   51.3    & 85.2 \\
    PBR   &   \underline{65.0}    &    \underline{47.2}   &   \underline{67.1}    & \underline{90.1} \\
    EIGML  &   \textbf{67.4{\large *}}    &   \textbf{50.0{\large *}}    &    \textbf{70.4{\large *}}   &  \textbf{92.1{\large *}} \\
\bottomrule
    \end{tabular}%
    \caption{Evaluation results on the DSTC10-MOD dataset, best results in \textbf{bold} while second with \underline{underline}. "\textbf{{\large *}}'' indicates the statistically significant improvements (i.e., two-sided t-test with $p<0.05$) over the best baseline.}
    \vspace{-15pt} 
  \label{tab:tabmod}%
\end{table}%

To provide a more comprehensive evaluation, we further assess our method on the DSTC10-MOD dataset. As shown in Table~\ref{tab:tabmod}, compared to the strongest baseline (65.0\% MAP and 47.2\% $R_{10}@1$), EIGML achieves absolute gains of 2.4\% in MAP and 2.8\% in $R_{10}@1$, respectively. These results further validate the effectiveness and robustness of our approach. By jointly modeling over the semantic content of the dialogue and incorporating both emotional and intentional cues, our method significantly improves sticker selection accuracy. Moreover, the results clearly demonstrate the strong generalization ability of EIGML.

\subsection{Effect of Emotion and Intention}

To further assess the effectiveness of incorporating emotional and intentional information beyond basic semantic alignment, we conduct targeted ablation studies to disentangle their individual contributions. Specifically, we independently remove the intention-related (w/o Intention) and emotion-related (w/o Emotion) modules from our model and evaluate the resulting changes in matching scores for positive sticker-text pairs. This setup allows us to analyze the distinct roles of emotion and intention modeling in enhancing SRS accuracy. Additionally, we compare our full model with a baseline Base, the model with only text-image semantic alignment and self-attention for fusion. These comparisons provide deeper insights into how emotion and intention contribute to understanding both stickers and dialogue.

\begin{figure}[t]
\setlength\abovecaptionskip{0.2\baselineskip}
\setlength\belowcaptionskip{0.2\baselineskip}
    \centering
    \includegraphics[width=1.0\linewidth]{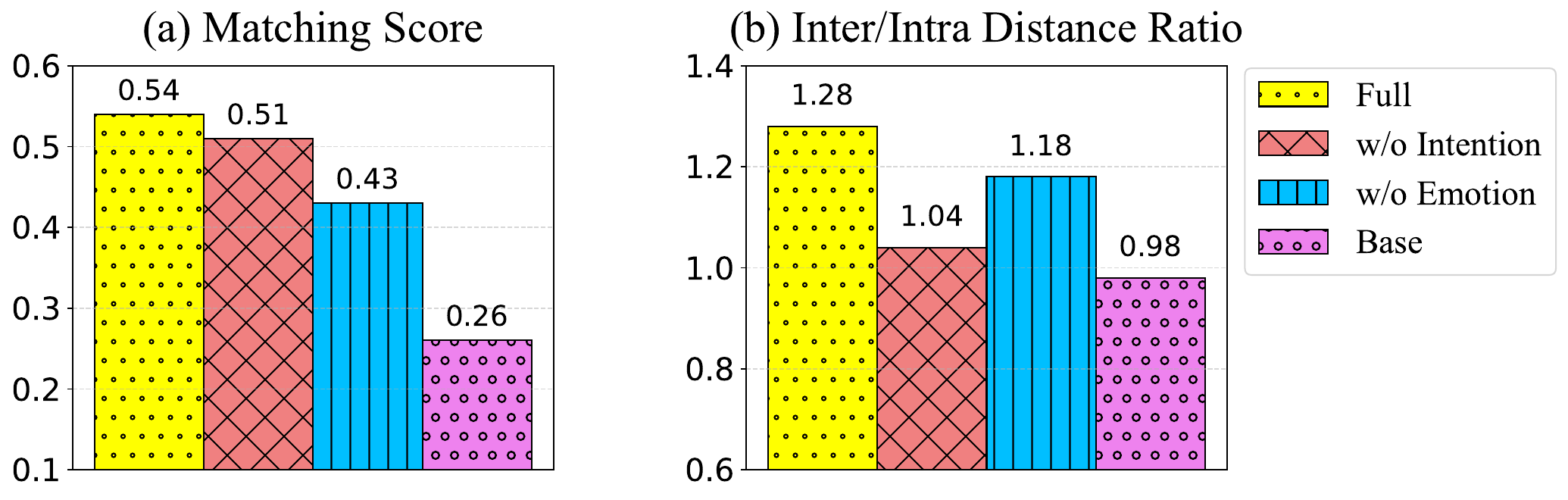}
    
    \caption{On StickerChat, matching scores and Inter/Intra Distance Ratios. Higher ratios reflect better class separation in the t-SNE space, while higher matching scores indicate stronger alignment between dialogues and correct stickers.}
    \label{fig:vis1}
    \vspace{-15pt} 
\end{figure}

The results are shown in Figure~\ref{fig:vis1}. We observe that our proposed EIGML model achieves an average matching score of 0.54 for correctly predicted positive sticker-text pairs. When the intention-related and emotion-related modules are removed, the scores drop to 0.51 and 0.43, respectively. In contrast, the base model only achieves a score of 0.26. These results clearly demonstrate the effectiveness of incorporating both intention and emotion modeling. The significant performance drop after removing either component underscores their complementary roles in understanding sticker emotion and intention, which are essential for mitigating mismatches when emotions and intentions are misaligned.

\begin{figure}[t]
\setlength\abovecaptionskip{0.2\baselineskip}
\setlength\belowcaptionskip{0.2\baselineskip}
    \centering
    \subfloat[Base]{
    		\includegraphics[scale=0.25]{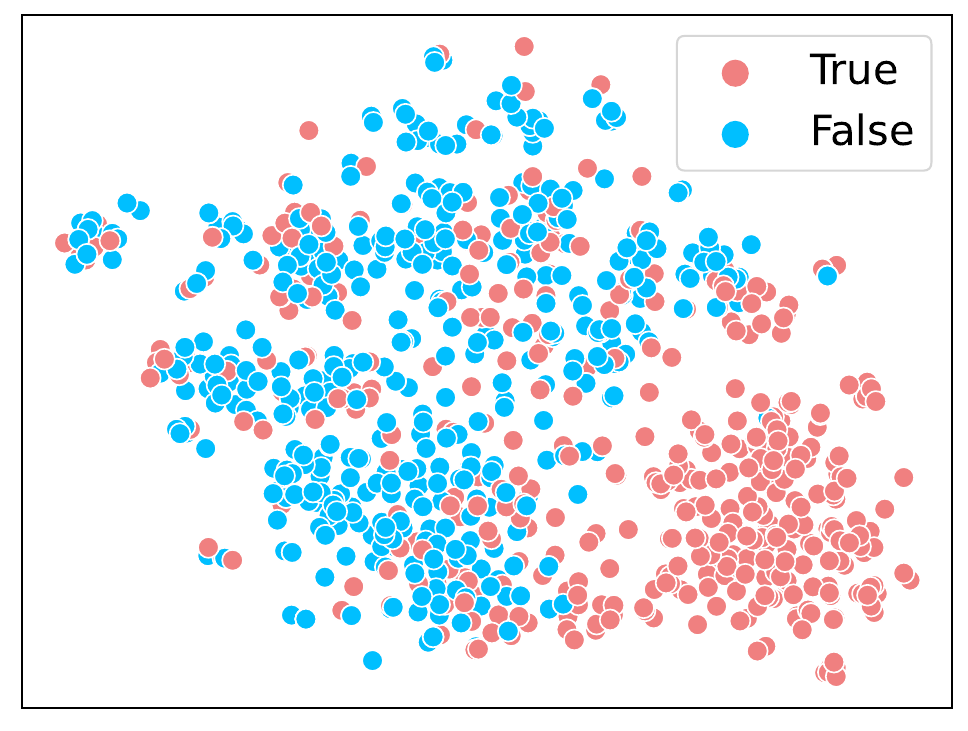}\label{fig:tsne_base}}
    \subfloat[w/o Emotion]{
    		\includegraphics[scale=0.25]{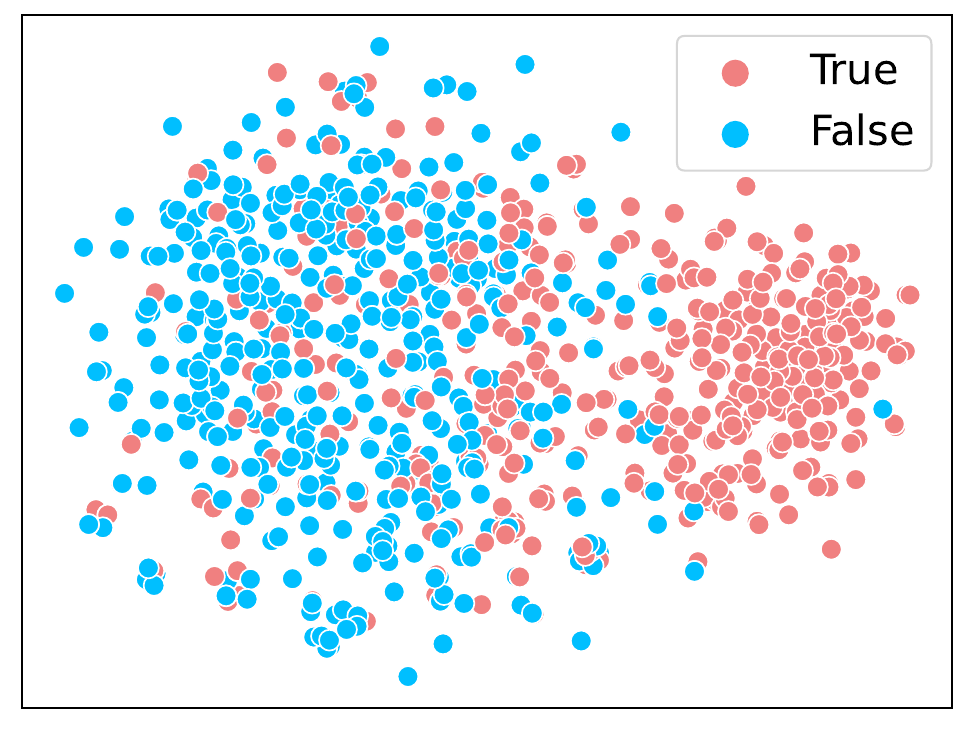}\label{fig:tsne_intention}}
    \\
    \subfloat[w/o Intention]{
    		\includegraphics[scale=0.25]{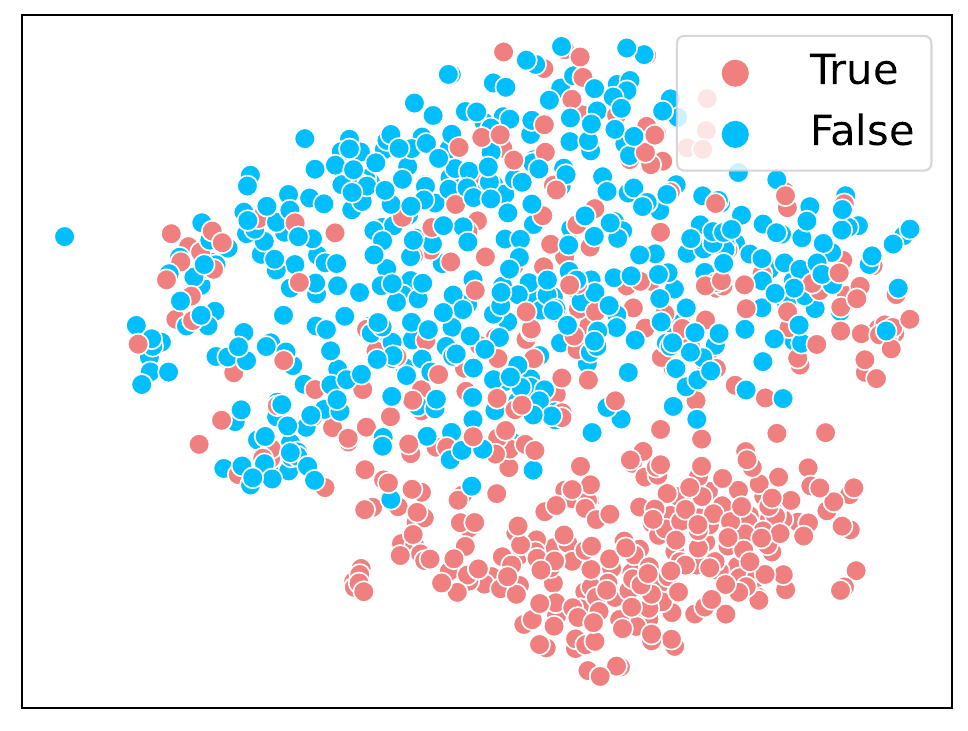}\label{fig:tsne_emotion}}
    \subfloat[EIGML]{
    		\includegraphics[scale=0.25]{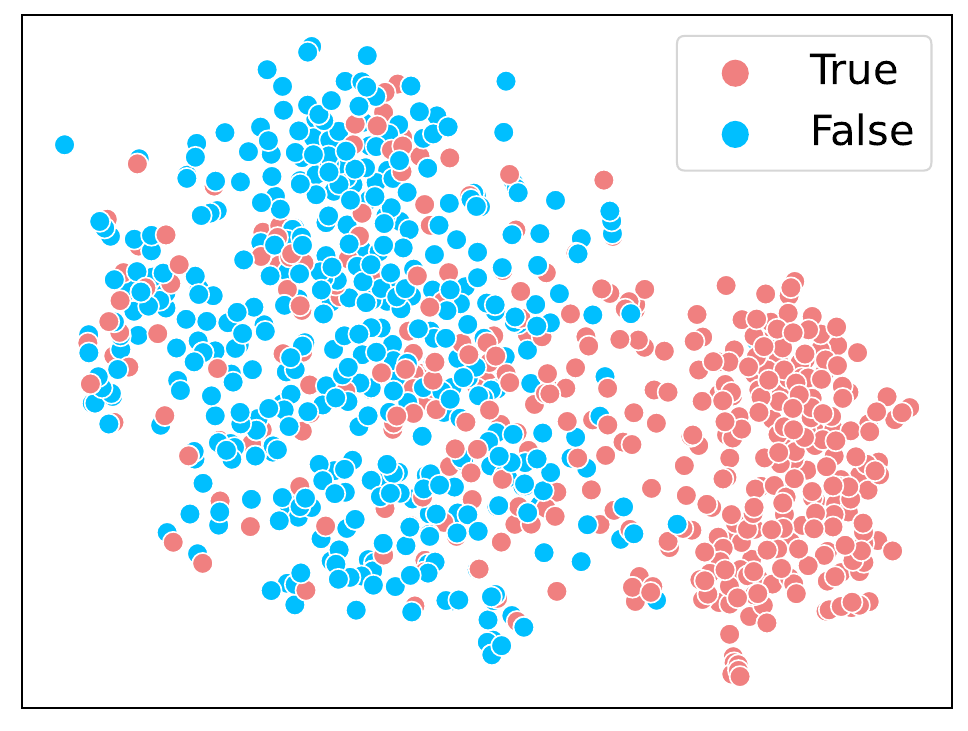}\label{fig:tsne_EIGML}}
    
    \caption{t-SNE visualization of 500 StickerChat examples using different methods. For each test instance, one ground-truth and one randomly sampled negative image (from nine distractors) are shown.}
    \label{fig:tsne}
    \vspace{-20pt}
\end{figure}

Furthermore, we perform a visualization analysis of sticker representations using t-distributed Stochastic Neighbor Embedding (t-SNE)~\cite{maaten2008visualizing}, as shown in Figure~\ref{fig:tsne}, and compare the Inter/Intra Cluster Distance Ratio of each method as shown in Figure~\ref{fig:vis1}, which measures the separability and compactness of learned representations. The results indicate that models lacking either emotional or intentional components tend to produce more scattered distributions, characterized by larger intra-class distances and smaller inter-class separations. In contrast, although the sticker features generated by EIGML also exhibit a certain degree of dispersion in the semantic space, they show a clearer distinction between correct and incorrect stickers. This observation suggests that jointly modeling both emotion and intention leads to more effective separation of relevant stickers in the feature space, thereby validating the importance of emotional and intentional information in understanding context and selecting appropriate stickers.

\subsection{Ablation Study}

To better assess the contributions of each key component in EIGML, we perform comprehensive ablation studies on the StickerChat dataset. EIGML comprises five core components: Inter-Modality Alignment (Inter), Intra-Modality Alignment (Intra), EIKS, IEGA, and SAMM. Furthermore, as image-text semantic alignment serves as the foundation of the SRS task, when ablating Inter, we only remove the emotion and intention alignment across modalities, while preserving the basic semantic alignment.

\begin{table}[t]
\setlength\abovecaptionskip{0.2\baselineskip}
\setlength\belowcaptionskip{-0.2\baselineskip}
  \centering
 
    \setlength{\tabcolsep}{8pt} 
    \scalebox{0.9}{
    \begin{tabular}{cccccc}
    \toprule
    Inter & Intra    & EIKS    & IEGA & SAMM  & MAP   \\
    \midrule
   
    $\surd$  & $\surd$     & $\surd$     & $\surd$     &   & 80.48  \\
    $\surd$     & $\surd$     & $\surd$     &       & $\surd$  & 80.58  \\
    $\surd$     & $\surd$     &       & $\surd$     & $\surd$  & 80.64  \\
    $\surd$     &       & $\surd$     & $\surd$     & $\surd$  & 80.99  \\
  &  $\surd$     &   $\surd$    & $\surd$     &   & 80.29  \\
    $\surd$     & $\surd$     & $\surd$     & $\surd$     & $\surd$& \textbf{81.15} \\
    \bottomrule
    \end{tabular}%
    }
     \caption{Ablation results on the StickerChat dataset. The meaning of $\surd$ is to include a submodule. Since SAMM is built upon Inter-modality emotion and intention alignment, it is also removed when Inter is ablated.}
     \vspace{-15pt} 
  \label{tab:ab1}%
\end{table}%

As shown in Table~\ref{tab:ab1}, each component of EIGML contributes to overall performance. Removing inter-modality alignment (Inter) and its associated SAMM module causes a notable performance drop (from 81.31\% to 80.29\%), highlighting the importance of emotion and intention alignment across modalities. Intra-modality alignment (Intra) also brings gains; its removal reduces performance to 80.99\%, suggesting consistency within each modality is vital for robust representation. Moreover, excluding either the EIKS module or IEGA leads to degradation (80.64\% and 80.58\%, respectively), showing the effectiveness of knowledge supervision and interaction-based guidance. Finally, disabling the similarity-based matching mechanism lowers performance to 80.48\%, confirming its complementary role in final prediction. Overall, these results validate the necessity of each component and the synergy of their integration.

\subsection{Hyper-parameter Analysis}

\begin{figure}[t]
\setlength\abovecaptionskip{0.2\baselineskip}
\setlength\belowcaptionskip{0.2\baselineskip}
    \centering
    \subfloat{
    		\includegraphics[scale=0.225]{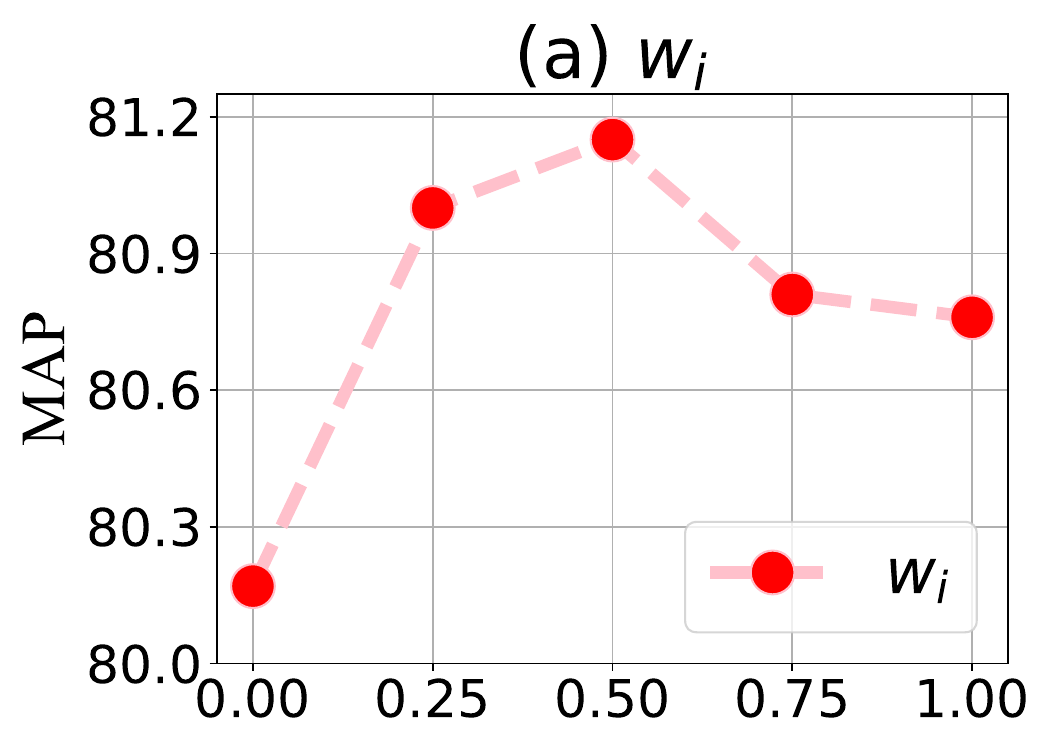}\label{fig:weight_i}}
    \subfloat{
    		\includegraphics[scale=0.225]{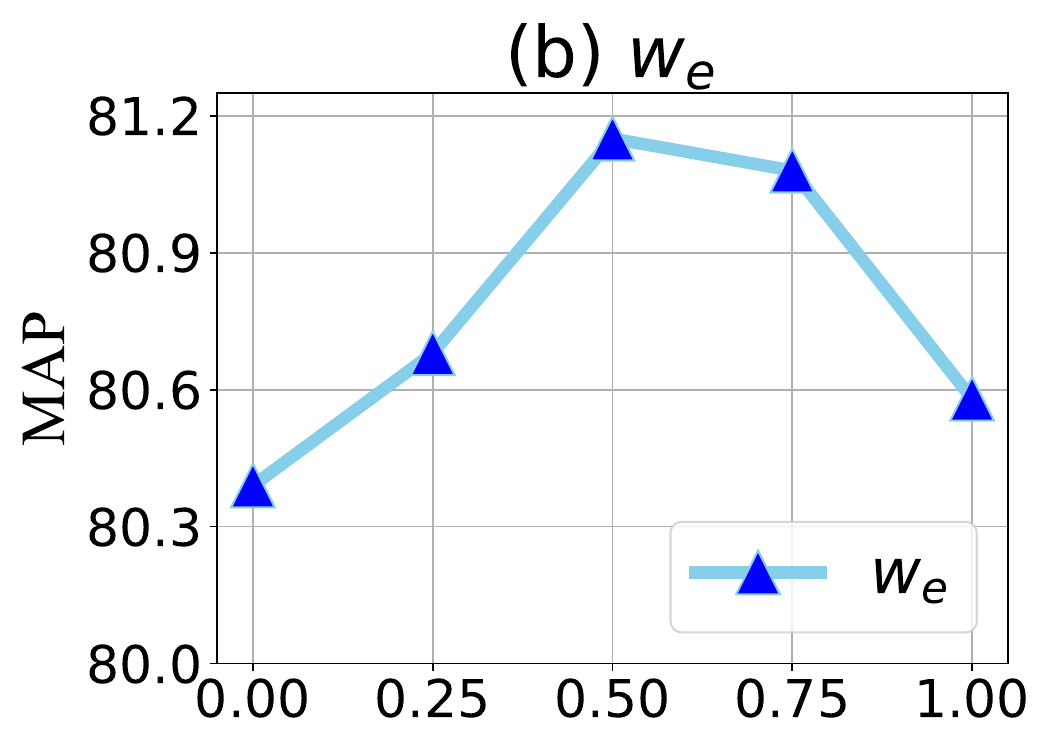}\label{fig:weight_e}}
    
    \caption{Hyper-parameter analysis on $w_{e}$ and $w_{i}$. When adjusting the $w_{e}$, $w_{i}$ is set as 0.5, and vice versa.}
    \vspace{-5pt}
    \label{fig:weight}
\end{figure}

We further investigate the hyperparameter used to control the weighting of emotion-related and intention-related losses, denoted as $w_e$ (including $w_e$, $w_e'$, $w_{e}^{t}$, and $w_{e}^{v}$) and $w_i$ (including $w_i$, $w_{i}^{t}$, and $w_{i}^{v}$), respectively. As shown in Figure~\ref{fig:weight}, we vary each weight from 0 to 1 in steps of 0.25, while keeping the other fixed at 0.5. The results reveal that performance, measured by MAP, is sensitive to both parameters. For $w_e$, the MAP peaks at 81.15\% when $w_e=0.5$, suggesting that moderate emphasis on emotion-related supervision yields optimal results. Similarly, the highest MAP for $w_i$ also appears at 0.5, further confirming that balanced incorporation of intention information is most effective. Notably, performance drops when either weight is set to 0 (i.e., the corresponding module is removed), indicating that both emotional and intentional signals are essential for maximizing model performance. These findings validate the importance of jointly modeling both aspects and show that their contributions are best realized when neither dominates.

\subsection{Computational Cost Analysis}

\begin{table}[t]
\setlength\abovecaptionskip{0.2\baselineskip}
\setlength\belowcaptionskip{-0.2\baselineskip}
  \centering
 
    \setlength{\tabcolsep}{8pt} 
    \scalebox{0.9}{
    \begin{tabular}{cccc}
    \toprule
    Model &	Training Time (it/s) &	Parameters (M) &	 FLOPs (T)\\
    \midrule
   
    PBR  & 0.5732     & 212.65    & 3.577     \\
    EIGML     & \textbf{0.5068}     & \textbf{196.64}     & \textbf{2.053}       \\
    \bottomrule
    \end{tabular}%
    }
     \caption{Computational cost comparison between EIGML and the strongest baseline PBR under the same setting with batch size 16, and a single A800 GPU. Best results in \textbf{bold}.}
  \label{tab:compution cost}%
  \vspace{-15pt} 
\end{table}%

To evaluate computational efficiency, we compare EIGML with the strongest baseline PBR under identical settings. As shown in Table~\ref{tab:compution cost}, EIGML reduces per-iteration training time, number of parameters, and FLOPS by 11.59$\%$, 7.53$\%$, and 42.61$\%$, respectively, achieving superior performance while substantially lowering computational overhead and validating its efficiency.


%% file: 5Conclusions.tex
\section{Conclusion}

In this paper, we propose Emotion and Intention Guided Multi-Modal Learning, a novel framework that overcomes the limitations of modeling emotion and intention separately in sticker response selection. By jointly integrating emotional and intentional cues, our framework enhances contextual understanding and selection accuracy. We introduce consistent intra- and inter-modality alignment strategies and progressively infuse contextual knowledge into the decision process. Extensive experiments on two benchmark datasets demonstrate the effectiveness and robustness of our approach. We hope this work inspires further research on multi-modal dialogue systems and sticker interactions, encouraging deeper exploration of fine-grained emotional and intentional modeling in real-world communication.